\documentclass[conference]{IEEEtran}
\IEEEoverridecommandlockouts
\usepackage{amsmath,amssymb,amsfonts}
\usepackage{amsthm}
\usepackage{algorithmic}
\usepackage{graphicx}
\usepackage{textcomp}
\usepackage{xcolor}
\usepackage{pgfplots}
\usepackage{tfrupee}
\usepackage{xfrac}
\usepackage{scalerel, stackengine}
\usepackage{bbm}
\usepackage{mathtools}
\usepackage{subcaption}
\usepackage{multirow}
\usepackage{float}
\usepackage{booktabs}
\usepackage{adjustbox}

\DeclarePairedDelimiter\abs{\lvert}{\rvert}

\stackMath
\newcommand\reallywidehat[1]{%
\savestack{\tmpbox}{\stretchto{%
  \scaleto{%
    \scalerel*[\widthof{\ensuremath{#1}}]{\kern.1pt\mathchar"0362\kern.1pt}%
    {\rule{0ex}{\textheight}}
  }{\textheight}%
}{2.4ex}}%
\stackon[-6.9pt]{#1}{\tmpbox}%
}
\pgfplotsset{compat=newest}

\pgfplotsset{
	tiny/.style={
		width=3cm,
		height=,
		legend style={font=\tiny},
		tick label style={font=\tiny},
		label style={font=\tiny},
		title style={font=\footnotesize},
		every axis title shift=0pt,
		max space between ticks=12,
		every mark/.append style={mark size=6},
		major tick length=0.1cm,
		minor tick length=0.066cm,
		every legend image post/.append style={scale=0.8},
	},
}

\newtheorem{problem}{Problem}
\newtheorem{theorem}{Theorem}[section]
\newtheorem{corollary}{Corollary}[theorem]
\newtheorem{lemma}{Lemma}[theorem]

\DeclareMathOperator*{\argmin}{arg\,min}
\newcommand{\beginparagraph}[1]{\noindent \textit{#1 }}
\newcommand{\A}{\textsc{SuperMAT}}

\newcommand{\Audience}{\textsc{AudienceCreation}}
\newcommand{\TIPAS}{\textsc{Tipas}}
\newcommand{\MF}{\textsc{MF}}
\newcommand{\Tophist}{\textsc{Top}}
\newcommand{\Toplatest}{\textsc{Top(45)}}
\newcommand{\BuyItAgain}{\textsc{BuyItAgain}}
\newcommand{\NewToCategory}{\textsc{NC}}
\newcommand{\OldToCategory}{\textsc{OC}}

\newcommand{\MKV}{\textsc{MKV}}
\newcommand{\LMKV}{\textsc{LMKV}}

\newcommand{\Real}{\mathbb{R}}

\newcommand{\PlotWeibull}[2]{
	\begin{tikzpicture}
		\begin{axis}[tiny, 
			axis x line=none,  
			axis y line=none,  
			minor tick num=1, 
			samples=100, 
			very thick, 
			domain=0:10]
			\addplot+ [black, no marks]
			{(#2 / #1) * (x / #1)^(#2 - 1) * exp(-(x / #1)^#2)};
		\end{axis}
	\end{tikzpicture}
}
\newcommand{\PlotQWeibull}[2]{
	\begin{tikzpicture}
		\begin{axis}[tiny, 
			axis x line=none,  
			axis y line=none,  
			minor tick num=1, 
			samples=100, 
			very thick, 
			domain=0:10]
			\addplot+ [black, no marks]
			{(#2 / #1) * (floor(x) / #1)^(#2 - 1) * exp(-(floor(x) / #1)^#2)};
		\end{axis}
	\end{tikzpicture}
}
\newcommand{\PlotMoW}{
	\begin{tikzpicture}
		\begin{axis}[tiny, 
			axis x line=none,  
			axis y line=none,  
			minor tick num=1, 
			samples=100, 
			very thick, 
			domain=0:30]
			\addplot+ [black, no marks]
			{0.5 * ((1.5 / 1) * (x / 1)^(1.5 - 1) * exp(-(x / 1)^1.5)) + 
			 0.5 * ((2.5 / 10) * (x / 10)^(2.5 - 1) * exp(-(x / 10)^2.5)};
		\end{axis}
	\end{tikzpicture}
}
\newcommand{\PlotQMoW}{
	\begin{tikzpicture}
		\begin{axis}[tiny, 
			axis x line=none,  
			axis y line=none,  
			minor tick num=1, 
			samples=100, 
			very thick, 
			domain=0:30]
			\addplot+ [black, no marks]
			{0.5 * ((1.5 / 1) * (floor(x) / 1)^(1.5 - 1) * exp(-(floor(x) / 1)^1.5)) + 
			 0.5 * ((2.5 / 10) * (floor(x) / 10)^(2.5 - 1) * exp(-(floor(x) / 10)^2.5)};
		\end{axis}
	\end{tikzpicture}
}

\ifCLASSINFOpdf
\else
\fi
\begin{document}
%
\title{Audience Creation for Consumables \\
     \Large Simple and Scalable Precision Merchandising for a Growing Marketplace}

\author{Shreyas S\IEEEauthorrefmark{1}\IEEEauthorrefmark{2},
\IEEEauthorblockN{Harsh Maheshwari\IEEEauthorrefmark{1}\IEEEauthorrefmark{2},
Avijit Saha\IEEEauthorrefmark{1}\IEEEauthorrefmark{2}, 
Samik Datta\IEEEauthorrefmark{1}\IEEEauthorrefmark{3}\IEEEauthorrefmark{4}, 
\thanks{\IEEEauthorrefmark{1} Equal Contribution, \IEEEauthorrefmark{4} Work done while at Flipkart}
\\
Shashank Jain\IEEEauthorrefmark{2}, 
Disha Makhija\IEEEauthorrefmark{5}\IEEEauthorrefmark{4},
Anuj Nagpal\IEEEauthorrefmark{2},
Sneha Shukla\IEEEauthorrefmark{2} and
Suyash S\IEEEauthorrefmark{2}}\\
\IEEEauthorblockA{\IEEEauthorrefmark{2}Flipkart Internet Private Limited\\ \{s.shreyas, harsh.maheshwari, avijit.saha \\ 
jain.shashank, anuj.nagpal, sneha.shukla, suyash.s\}@flipkart.com, \\
\IEEEauthorrefmark{3}Amazon \\ 
\{datta.samik@gmail.com\},\\ \IEEEauthorrefmark{5}The University of Texas at Austin \\ \{dishadmakhija@gmail.com\}
}}


\maketitle

\begin{abstract}

Consumable categories, such as grocery and fast-moving consumer goods, are quintessential to the growth of e-commerce marketplaces in developing countries. In this work, we present the design and implementation of a precision merchandising system, which creates audience sets from over $10$ million consumers and is deployed at Flipkart {\em Supermart}, one of the largest online grocery stores in India. We employ temporal point process to model the latent periodicity and mutual-excitation in the purchase dynamics of consumables. Further, we develop a likelihood-free estimation procedure that is robust against data sparsity, censure and noise typical of a growing marketplace. Lastly, we scale the inference by quantizing the triggering kernels and exploiting sparse matrix-vector multiplication primitive available on a commercial distributed linear algebra backend. In operation spanning more than a year, we have witnessed a consistent increase in click-through rate in the range of $25\text{--}70\%$ for banner-based merchandising in the storefront, and in the range of $12\text{--}26\%$ for push notification-based campaigns.



%
%
\end{abstract}


%
\IEEEpeerreviewmaketitle


 



\section{Introduction} \label{Introduction}

\noindent While the e-commerce market in the West is nearing saturation, the emerging economies are poised to become the next frontier of the global e-commerce boom. E-commerce in India -- one of the largest economies of the developing world -- is projected to grow at a cumulative annual growth rate of $30\%$, reaching \textdollar$84$ billion by the year $2021$, which, sadly, would still represent under $10\%$ of its retail market.

\noindent Given the immense growth opportunity and the burgeoning market, customer growth and retention metrics -- such as monthly active customer and transactions per customer -- are of supreme interest to the e-commerce players in these markets. To improve retention, one of the strategies that these players have adopted is to expand into new categories for consumables:  such as grocery and fast-moving consumer goods (FMCG).


\noindent Accordingly, the merchandising of consumables has become an important problem. For merchandising, the category managers target the consumers with deals and discounts to meet monthly sales targets. The merchandising at Flipkart follows a two-stage design: 1) Audience Creation: a sufficiently large audience set is created for each category and 2) Audience Manager: following audience creation, the deals/discounts available in a category becomes eligible to be shown to the category specific audience. However, a consumer can be associated with multiple audience sets. Therefore, the audience manager further refines these consumer specific deals/discounts. 








\noindent In this work, we focus on the audience creation step. Importantly, every consumer can not belong to each audience set because the audience manager can only deal with a handful of deals/discounts due to the latency requirement. Hence, intuitively, the audience set for a category should contain the consumers who have higher chance of conversion from the category. This warrants an algorithmic approach to audience creation, rather than relying on the tradition human-in-the-loop predicate-based approach.

\noindent We implement this by incorporating the latent periodicity (e.g., consumers purchase $50$ tea bags every month) and cross-excitation (e.g., purchase of beverages excites the purchase of snacks) in the purchase dynamics into the audience creation process by embracing the framework of temporal point process to model the probability of purchase.

\noindent Growing marketplaces bring about further nuances: sparsity in data due to marketplace fragmentation and frequent category launches, as well as, noise injected by rapid purchases made by abusers. We adjust our estimation procedure accordingly by using Likelihood Free Estimation i.e., deconstructing it into two phases: estimation of the triggering kernel which captures the temporal purchase dynamics and estimation of the latent network which models the cross excitation. This aides us to compensate for the aforementioned nuances.

\noindent We fulfil the need of scale and frequent inference updates by quantising the temporal point process and casting the convolution operation in the inference path as distributed sparse matrix-vector multiplication (SpMV) readily available in linear algebra backends.

\noindent We now enumerate the contributions:
\begin{enumerate}
	\item We embrace the temporal point process framework to model the underlying periodicity and mutual-excitation in the purchase dynamics of the consumables.
	\item We deconstruct the estimation procedure to adapt to data sparsity and noise caused by frequent category launches, fragmented marketplace, frequent purchases by abusers and aberrations created by heavy incentivisation-led purchases.
	\item We scale the inference by quantising the temporal point process and casting the convolution operation as sparse matrix-vector multiplications atop distributed linear algebra backends.
	\item The audience creation system is deployed at Flipkart Supermart for last 12 months and scales to $\sim 10$ million consumers and $\sim 200$ categories and brings about a consistent increase in clickthrough rate in the range of $25-70\%$ for banner-based merchandising in the storefront, and in the range of $12-26\%$ for push notification-based campaigns.
\end{enumerate}

Rest of the paper is organised as follows. In Section \ref{Problem}, we describe the problem formulation. The related work is reviewed in Section \ref{Related}. Section \ref{Background}, \ref{Modeling}, \ref{Estimation} and \ref{Inference} presents the background on Temporal point process, modeling details, parameter estimation, and inference, respectively. The system design and implementation related details are highlighted in Section \ref{System}. We provide the offline and online experimental results in Section \ref{Experiments}. Finally, we conclude our work in Section \ref{Conclusion}.

\section{Problem Setting} \label{Problem}
\subsection{Background}
\textit{Supermart.} India's leading e-commerce player Flipkart's online grocery store, was launched in $2017$ in Bangalore and has since been expanding operations across the country. Supermart now serves millions of customers in hundreds of consumable categories across several cities in India.

\begin{figure}
    \centering
    \includegraphics[width=0.45\columnwidth, height=6cm]{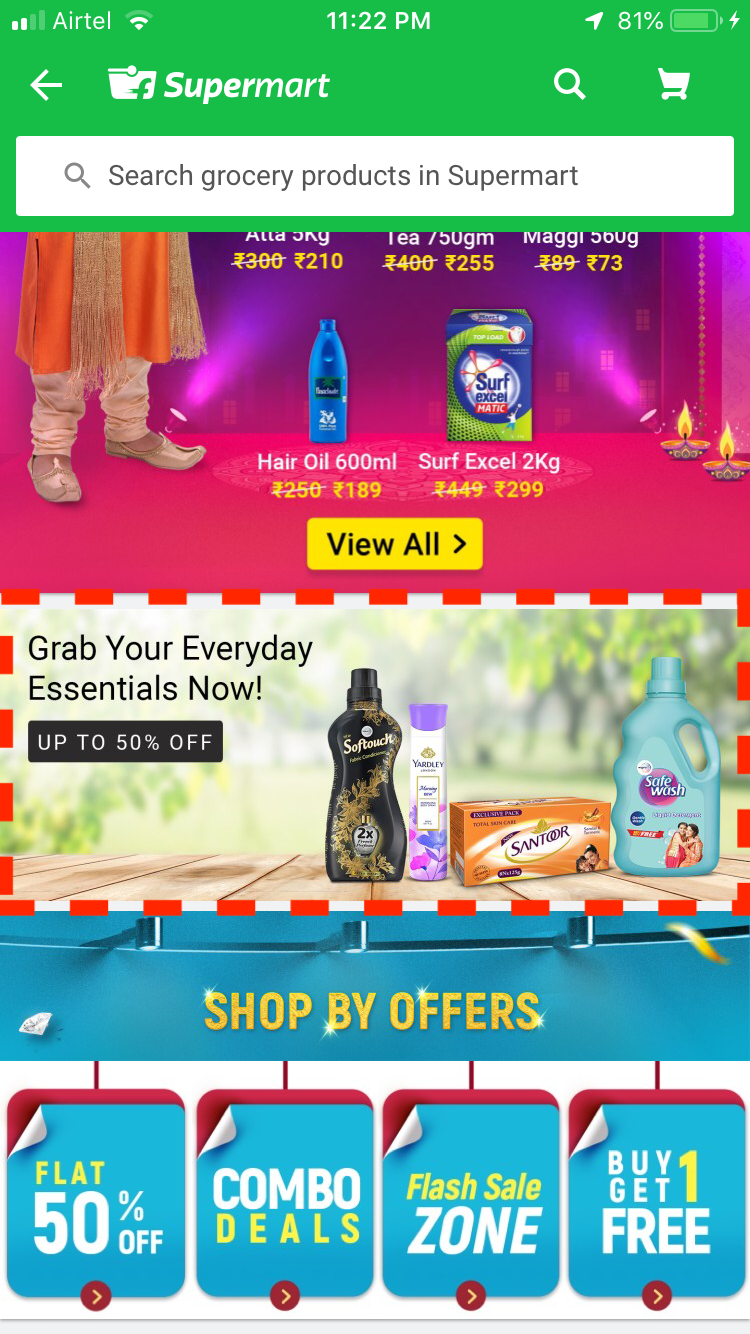}
    \caption{A banner in Supermart homepage merchandising everyday essentials at up to $50\%$ discount to a target audience.}
    \label{fig:supermart_banner}
\end{figure}

\noindent At this phase of growth, merchandising, where category managers spend an allotted budget in the form of deals and discounts to meet monthly sales targets, acts as the foremost sales channel. Aggressive sales tactics –- such as \rupee$1$ deals and $50 \%$ discounts (see  Fig. ~\ref{fig:supermart_banner} for an illustration, where a \textit{banner} merchandises everyday essentials on up to $50 \%$ discount) -- are deployed frequently to drive up the traffic and the sales. Because of the strict latency requirement of downstream tasks, the audience creation step warrants a {\em precision merchandising} strategy, wherein only a subset of the users are made eligible for each of the deal/discount while maximizing the conversion.

\noindent The traditional audience creation process for precision merchandising calls for defining the target audience with predicates: e.g., the target audience for beverages are those who have purchased beverages in the past or have viewed them recently. However, the growing nature of the business makes this human-in-the-loop process untenable. Newly launched categories would not have enough past purchases and such predicates would not furnish the requisite {\em reach} (equal to the audience size). While this can be mended by adding new clauses to the predicate that include purchases from {\em related} categories, it is not reasonable to expect each of merchandisers to share the same view of relatedness of the categories, thus limiting the efficacy of this strategy.

\noindent Additionally, the audience creation process for consumables would need to cater to repeat purchases: there is a certain periodicity in purchases of consumables that has to be embodied into the predicates. While the periodicity can, in theory, be estimated from the purchase logs, the estimation procedure is fraught with nuances: a growing market such as India has several competing online grocery stores, resulting in a fragmentation of the purchase logs. Furthermore, there are abusers who purchase frequently in order to re-sell it later, creating aberrations in the purchase log. This warrants a robust estimation procedure and strongly discourages the tradition human-in-the-loop practise of audience creation.
\subsection{Product Desiderata}
\noindent Based on the aforementioned observations, we are now in a position to articulate the product desiderata:
\begin{itemize}
	\item \textbf{Precision.} We need to match the incentives with users with a high probability of conversion, instead of broadcasting it.
	\item \textbf{Reach} The size of the audience set should conform to certain lower limit and certainly should be larger in size compare to the audience set created by the traditional approaches.
	\item \textbf{Simplicity.} Given the inadequacy of the traditional human-in-the-loop audience creation process, we need to systemically embody the notions of relatedness of categories and repeat purchase periodicities into the audience creation process, rather than exposing these complexities to the category managers.
    \item \textbf{Robustness.} The estimation of the category relatedness needs to handle the long-tail of category popularities (e.g., staples and beverages are heavily purchased, whereas certain snacks are less popular). Purchase periodicity estimation also has to be robust against the aberrations created by the abusers and the fragmented market.
	\item \textbf{Scalability.} Given the scale of operation ($\sim 10$ million users and $\sim 200$ categories -- ever expanding) and the dynamic nature of the business (frequent new category launches), the audience creation process needs to run frequently and need to scale efficiently.
\end{itemize}

\subsection{Problem Formulation}

\noindent We begin with recounting the relevant concepts from e-commerce, and, along the way, set up the notation for the remainder of the paper. Let $\mathcal{U}$ denote the universe of the users in the e-commerce platform and let $\mathcal{I}$ denote the set of items in its catalogue. Each item, $i \in \mathcal{I}$ further belongs to the category, $c_i \in \mathcal{C}$, with $\mathcal{C}$ being the set of categories.

\noindent The users' interactions with the items' in the platform during the observation window $[0, t)$ are persisted in the form of a collection of \textit{behavioural logs}, $\mathcal{H}_u(t), \forall u \in \mathcal{U}$. The  behavioural log for user $u$ is time-ordered collection of tuples: $\mathcal{H}_u(t) = \{ (i_n, t_n) \mid t_n < t\}_{n=1}^{N_u(t)}$, where each tuple represents a purchase done by user $u$ on item $i_n$ at time instance $t_n$. And $N_u(t)$ denotes the number of interactions recorded for user $u$ until time $t$.

\noindent With this background, we are now in a position to formally state the problem -- \Audience \space -- we set out to solve in this paper.

\begin{problem}[\Audience]
	At any given time $t$, for each category, $c \in \mathcal{C}$, compute a total order on the universe of users, $(\mathcal{U}, \succ_c)$, that allows us to rank and select the top as the audience, depending on its reach requirement. The total order sorts users based on the estimate of the probability of purchase from category $c$ in a stipulated time window $[t, t + \delta)$ \footnote{We set $\delta = 9$ days in our experiments.}.
\end{problem}
\noindent We, next, define the related problem of purchase probability estimation:
\begin{problem}[\A]
	Given $\mathcal{H}_u(t), \forall u \in \mathcal{U}$, estimate the set of purchase probabilities, $p_{u, c}(t), \forall u \in \mathcal{U}, \forall c \in \mathcal{C}$, that signifies purchases ($u$ purchasing an item from category $c$) in a stipulated time window $[t, t + \delta)$.
\end{problem}

\section{Related Work} \label{Related}

\beginparagraph{Recommendation System.} The central goal of a recommendation system -- "recommending" a set of items to each user that they are likely to purchase -- is similar in spirit, except that in merchandising, we recommend a set of users for each item. Matrix factorisation and its variants (see~\cite{mf, imf}) have long dominated this field, but they ignore the underlying temporal dynamics of purchases.~\cite{fpmc} refines the approach by modelling the sequence explicitly, but ignores the temporal information.~\cite{wei} further refines the approach by modeling the individuals' inter-purchase intervals, but do not model the category-specific replenishment cycles and cross-category excitation.

\beginparagraph{Time-sensitive Recommendation.} \cite{rahul} employs a Poisson process to model the probability of purchase in case of consumables, and is closest to our problem. However, it does not degrade gracefully with data sparsity, a major design consideration for a growing marketplace. {\TIPAS} ~\cite{jure}, which is a state-of-the-art method, extends the dynamics to model short-term excitation and long-term periodicity. It subsumes most of the kernel design choices for consumables in an e-commerce setting, however fails to model the purchase dynamics idiosyncratic to Supermart as we shall observe in Section \ref{Modeling}. Addtionally, {\TIPAS} has been comprehensively bench-marked and proven to be superior than deep learning methods. ~\cite{alex} further extends the dynamics to be non-linear, but scales poorly with the number of users.

\beginparagraph{Precision Marketing.} The study of marketing in the computer science community spans at least two decades. In one of the earliest works on the subject, \cite{kitts} exploits co-occurrence in purchase logs to "cross-sell" items;~\cite{marketingbudget} examines the problem of optimal allocation of marketing budget to a portfolio of campaigns;~\cite{precisionmarketing} employs causal inference to estimate the effectiveness of campaigns in order to further refine budget allocation. However, none of the works focus on the repeat purchase behaviour that dominate consumables. Nor is optimal budget allocation the subject of the present study.

\noindent In this work, we rely on temporal point process (see~\cite{tpp} for an exposition) to express the purchase dynamics, and let it's literature to inform the design of a likelihood-free estimation procedure. For inference, we rely on quantization (\cite{scott}), and deploy  computational tricks such as "lowering" a convolution into matrix multiplication (\cite{chetlur2014cudnn}).

\section{PRELIMINARIES} \label{Background}
In this section, we discuss the technical details of temporal point process framework which we shall use to model the \Audience.

\subsection{Hawkes Point Process}
\noindent  We begin with an account of the Hawkes process~\cite{lecture}, a temporal point process, popularly employed to model a sequence of events - $\{ t_i \}_{i=1}^{n}$, where $t_i \in \mathbb{R}$, $t_i < t_j$ when $i < j$, and $n$ is the number of events. The Hawkes process is succinctly described with the following equation termed as close conditional intensity function:
\begin{equation} \label{equ:hawkes-process}
	\lambda_{u,c}^*(t) = \mu_{u,c}(t) + \sum_{c' \in \mathcal{C}} \beta_{c,c'} \int_0^t \kappa(t - \tau; \theta_{c,c'}) dN_{u,c'}(\tau)
\end{equation}
where $dN_{u,c}(t)$ denotes the number of events created by $u$ on category $c$ in the infinitesimal interval $[t, t + dt)$. Intuitively, $\lambda_{u,c}^*$ captures the probability of occurrence of an event in the infinitesimal interval $[t, t + dt)$. Below, we describe the individual components in detail.

\beginparagraph{Time-varying Intensity.} $\mu_{u,c}(t) \geq 0$ captures a time-varying intensity which for the purpose of the present work, $\mu_{u,c}(t) = \mu_c$. Popular categories, such as \textit{Sona Massori Rice} (a staple in several Indian states), intuitively, possesses larger values of $\mu_c$.

\beginparagraph{Latent Network.} $\beta_{c,c'} \geq 0$ denotes the (latent) network amongst the categories, $\mathcal{C}$. In particular, $\beta_{c,c'}$ denotes the influence of an event happening on category $c'$ on the probability of occurrence of an event in category $c$. As an example, a purchase occurring in 
\textit{Noodles} arguably raises the probability of another purchase in 
\textit{Ketchups}, due to the complementary nature of these two categories.

\beginparagraph{Triggering Kernel.} The family of functions $\kappa(\cdot; \theta_{c,c'}): \mathbb{R}_+ \mapsto \mathbb{R}_+$, are known as \textit{triggering kernels}. Intuitively, it captures the influence of an old event ($t - \tau$ being its age) at the present time. A rich assortment of triggering kernels exist in the literature like Hawkes or Exponential~\cite{leman} to model self-excitations, and more recently Weibull~\cite{jure} to model periodic sequences. We refer to Table \ref{tab:compendium} for further details.
\vspace{-0.20cm}
\begin{table*}
  \caption{A Compendium of Triggering Kernels} \label{tab:compendium}
  \begin{tabular}{rccll}
      \toprule 
      Kernel & Parametric Form ($\kappa(\cdot; \Theta)$) & Parameter ($\Theta$) & Shape & Shape After Quantisation \\
      \midrule
      \textsc{Hawkes or Exponential} & $\exp \Big(- \frac{t - \tau}{\omega} \Big)$ & $\omega$ & \parbox[c]{3.5cm}{\PlotWeibull{1.5}{1}} & \parbox[c]{3.5cm}{\PlotQWeibull{1.5}{1}} \\ 
      \textsc{Weibull} & $\frac{k}{\lambda} \Big( \frac{t - \tau}{\lambda} \Big)^{k - 1} \times \exp \Big(- \frac{t - \tau}{\lambda} \Big)^k$ & $\lambda, k$ & \parbox[c]{3.5cm}{\PlotWeibull{1}{1.5}} & \parbox[c]{3.5cm}{\PlotQWeibull{1}{1.5}} \\
      \textsc{MoW} & $\sum_{i=1}^{K} b_i \kappa_{W}(t - \tau; \lambda_i, k_i)$ & $\Bigg\{ \lambda_i, k_i, b_i \Bigg\}_{i=1}^{K}$ & \parbox[c]{3.5cm}{\PlotMoW} & \parbox[c]{3.5cm}{\PlotQMoW} \\
      \bottomrule
      \end{tabular}
\end{table*}

\subsection{Modeling Purchase Dynamics}
Here, we study the typical purchase behaviours observed in a e-commerce setting and their modelling with the temporal point process. Typically, the distribution of the inter-purchase times of the customers across various categories guides the modeling choices.

\beginparagraph{Excitation.} 
Usually, users buy products from different or same categories, together or within a short span of few days as evidenced in the customers' shopping history. This behaviour is typically modeled with Hawkes Process with an exponential kernel ~\cite{leman}.

\begin{itemize}
	\item \textsc{Hawkes}: assumes the form: $\kappa(t - \tau; \omega_{c,c'}) = \exp \left(- \frac{t - \tau}{\omega_{c,c'}} \right)$, where the influence drops monotonically at the rate controlled by the scale parameter, $\omega_{c,c'}$ to model the short span of the excitation which purchase in $c'$ triggers in another complementary category, $c$. The $\beta_{c,c'}$ models the strength of the relationship between the categories $c$ and $c'$ .
\end{itemize}

\begin{figure}[htp]
  \centering
  \begin{subfigure}[b]{0.4\textwidth}
    \includegraphics[width=\textwidth]{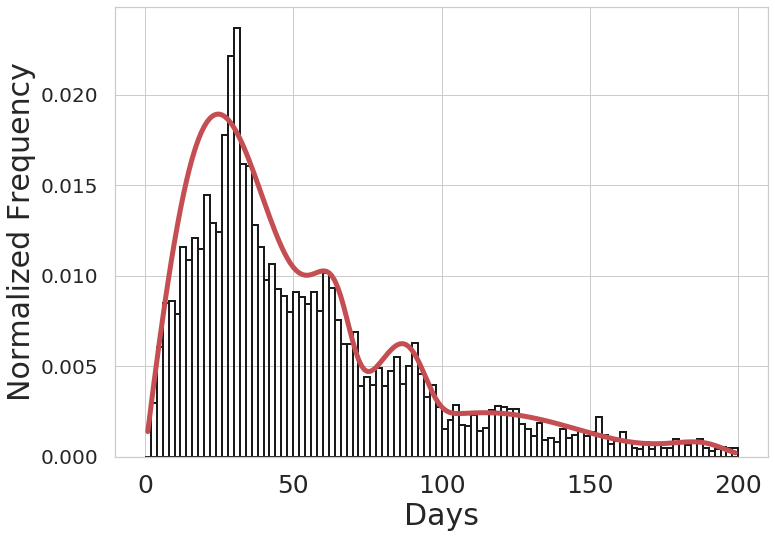}
  \end{subfigure}
  \hfill
  \caption{Histogram of inter purchase times for Sugar along with \textsc{MoW} fit.}
  \label{fig:mow_fit}
\end{figure}

\beginparagraph{Periodicity.}
The purchases from the same category exhibit a different behaviour where post the initial decay subsequent peak is encountered in the distribution of inter-purchase times which has been modeled with a Weibull Kernel ~\cite{jure}.
\begin{itemize}

\item \textsc{Weibull}: models a delayed influence and assumes the form: $\kappa(t - \tau; \lambda_{c,c'}, k_{c,c'}) = \frac{k_{c,c'}}{\lambda_{c,c'}} \left( \frac{t - \tau}{\lambda_{c,c'}} \right)^{k_{c,c'} - 1} \times \exp \left(- \frac{t - \tau}{\lambda_{c,c'}} \right)^{k_{c,c'}}$. As before, $\lambda_{c,c'}$ serves as the scale parameter, whereas, $k_{c,c'}$ controls the shape (the location of the peak influence). Evidently, an Weibull kernel can capture the periodicity in purchase with $k_{c,c}$ controlling the inter-purchase period (roughly $1$ month).
\end{itemize}

\section{Modeling} \label{Modeling}
\noindent In this section, we lay down the modelling methodology and motivate the development of \A, which lead to the solution of this paper's main concern \Audience.

\subsection{Modeling Purchase Dynamics at Supermart}
Here, we dwell upon the idiosyncrasies in the purchase behaviours at the Flipkart Supermart marketplace and incorporate them into the design of the conditional intensity function that underlies \A. Concretely,  we study the distribution of the inter-purchase times of our customers across various categories and use them to guide the modelling choices. We also introduce a novel triggering kernel, hitherto not studied in the literature i.e, Mixture of Weibulls.

\beginparagraph{Cross excitation.}  Typically, users buy complementary products of different categories, such as \textit{Noodles} and \textit{Ketchup}, together or within a short span of few days and we resort to model it with exponential kernel as already found in the literature.

\beginparagraph{Periodicity with Attenuation.} 
We encounter periodicity of roughly 30 days on analysing inter purchases within the same category, a consequence of the regular monthly purchasing behaviour of the {\A} customers. In figure \ref{fig:mow_fit} the category \textit{Sugar} 
shows a peak around 30 days with a high mass concentrated around it. However a curious behaviour observed across several categories of {\A} is of periodicity with attenuation i.e., distribution of times of repeat purchases for the same category exhibits multiple peaks repeating with a period of multiples of 30 days but with successive peaks of decreasing intensity. This phenomenon has not been observed so far in the study of repeat purchase of consumables. Nonetheless, the attenuated peaks are clearly intuitive as some people may be bi-monthly buyers, or factors like presence of competitors or discounts/offers, which results in people buying more and hence delaying their next purchase. Figure \ref{fig:mow_fit} clearly showcases the periodic peaks that are successively attenuated and a Weibull kernel alone fall short of capturing this behaviour, so we introduce a novel kernel.
\begin{itemize}
\item Mixture of Weibulls \textsc{MoW}: models both the delayed influence/periodicity with attenuation and assumes the form  $\kappa(t - \tau; \{ \lambda_i, k_i, b_i \}_{i=1}^{K}) = \sum_{i=1}^{K} b_i \kappa_{W}(t - \tau; \lambda_i, k_i)$ where $\kappa_{W}$ is the \textsc{Weibull} kernel. We note that \textsc{Weibull} and \textsc{Hawkes} kernels are special cases of \textsc{MoW}, which are seen by setting $K =1$ for \textsc{Weibull} and additionally setting $k_{c,c'} = 1$ for \textsc{Hawkes}.
\end{itemize}
\beginparagraph{\textsc{MoW} is a Conditional Intensity Function.} We conclude this section by reviewing a characterisation of a conditional intensity function (CIF) to derive a closure property that will assist us in showing that \textsc{MoW} is a valid CIF.
\begin{theorem}(Folklore)
	A conditional intensity function $\lambda^*(t)$ uniquely defines a temporal point process if it satisfies the following properties for all realisations $\{t_i\}_{i = 1}^{n}$ and for all $t > t_n$:
	\begin{enumerate}
		\item $\lambda^*(t)$ is non-negative and integrable on any interval starting at $t_n$, and
		\item $\int_{t_n}^{T} \lambda^*(t) dt \xrightarrow[T \to \infty]{} \infty$
	\end{enumerate}
\end{theorem}
\noindent We omit the proof for brevity and refer the interested reader to \cite{lecture}.
\begin{lemma}(Closure)
	A conic combination of a finite collection of conditional intensity functions, $\{ \lambda^*_{i}(t) \}_{i=1}^{K}$, is also a conditional intensity function.
\end{lemma}
\begin{corollary}(MoW)
	The \textsc{MoW} triggering kernel, defined as a conic combination of a finite collection of \textsc{Weibull} triggering kernels, $\{ \kappa_{W}(t - \tau; \lambda_i, k_i) \}_{i=1}^{K}$, leads to a well-defined conditional intensity function.
\end{corollary}

\subsection{Dynamics of \A}
\beginparagraph{Quantisation.} In an attempt to scale the inference for \A, we depart from the continuous time domain, and, instead, work with the (discrete) ticks of the wall-clock. In particular, we use $\delta$ to denote the grain of time (we set it to $9$ days in this work), and partition the interval $[0, t)$ into $\lceil \frac{t}{\delta} \rceil$ subintervals indexed by $s$. In a similar vein, we quantise the triggering kernels: inside each subinterval, $[s \delta, (s + 1) \delta)$, the \textit{quantised triggering kernel}, $\kappa_{\theta_{c, c'}}[s]$, now assumes a constant value that is equal to $\kappa(s \delta; \theta_{c, c'})$. With a slight abuse of notation, we continue to denote the piece-wise constant quantised triggering kernel with $\kappa_{\theta_{c, c'}}[\cdot]$ (see table~\ref{tab:compendium} for an illustration). The subsequent quantisation of the counting process follows easily: we denote the number of events occurring during the $s^\text{th}$ subinterval with  $N_{u, c'}[s]$, where $N_{u, c'}[s] = \int_{s \delta}^{(s + 1) \delta} dN_{u, c'}(\tau)$. Lastly, in a step reminiscent of the numerical quadrature, we approximate the definite integral in ~\eqref{equ:hawkes-process} as follows:
\begin{equation}\label{eq:quantised}
	\int_0^t \kappa(t - \tau; \theta_{c, c'}) dN_{u, c'}(\tau) \approx \sum_{s=0}^{\lceil \frac{t}{\delta} \rceil - 1} \kappa[\lceil \frac{t}{\delta} \rceil - 1 - s; \theta_{c, c'}] N_{u, c'}[s]
\end{equation} 

\beginparagraph{Convolution.} Clearly \eqref{eq:quantised} can be expressed as convolution operation $\star$ and can be written more succinctly as 
\begin{equation}\label{eq:convolution}
	\int_0^t \kappa(t - \tau; \theta_{c, c'}) dN_{u, c'}(\tau)
	 = \left( \kappa[;\theta_{c, c'}] \star N_{u, c'} \right) [\lceil \frac{t}{\delta} \rceil - 1]
\end{equation}
\noindent Apart from clarity, by expressing as convolution operation we leverage scaling the inference of \A as discussed in Section $5$.

\beginparagraph{CIF for \A~Dynamics.}
We are now in a position to state the intensity function of \A by utilizing \eqref{eq:convolution}
\begin{equation} \label{equ:SuperMAT}
	\lambda_{u,c}^*(t) = \mu^0_{c} + \sum_{c' \in \mathcal{C}} \beta_{c,c'} \left( \kappa[; \{ \lambda_i, k_i, b_i \}_{i=1}^{K}] \star N_{u, c'} \right) [\lceil \frac{t}{\delta} \rceil - 1]
\end{equation}
where $\kappa[; \{ \lambda_i, k_i, b_i \}_{i=1}^{K}]$ is the quantized version of the \textsc{MoW} kernel from Table \ref{tab:compendium}. We model the base intensity $\mu^0_{c}$ as purely the function of category $c$.

\section{Estimation} \label{Estimation}
At this point, we note that we have a mix of organic and promotional transactions in the dataset. Similarly, the customer-base comprise of genuine consumers, as well as \textit{re-sellers} -- the customers (abusers) who buy from E-commerce marketplaces only to sell at their retail outlets, thus commanding an arbitrage in the process. Moreover, owing to the frequent expansions in the categories, the category popularity distribution follows a Pareto, leaving insufficient data for a monolithic maximum likelihood estimation algorithm to work.

\beginparagraph{Likelihood Free Estimation.} In order to address the aforementioned idiosyncrasies, we deviate from the monolithic MLE estimation procedure that frequents the state-of-the-art literature, and, instead, deconstruct the estimation procedure into stages that estimate the latent network and the triggering kernel parameters separately. This choice allows us to parallelise the estimation procedure, as well as to modularise it -- as an example, one can choose an estimation procedure for the latent network from an available assortment of $3$. We detail the proposed algorithm below where we explicitly filter out the noise injected by \textit{re-sellers} and frequent, heavily discounted category specific launches. 

\subsection{Base Intensity Estimation}
Extending our notation, we let $N_{u,c}(t)$ denote the number of times user $u \in \mathcal{U}$ has purchases from category $c \in \mathcal{C}$ until time $t$. Similarly, we let $N_c(t) = \sum_{u \in \mathcal{U}} N_{u,c}(t)$ denote the total number of purchases from category $c$. The base intensities are estimated as follows, where $T$ denotes the span of the training dataset (measured in days):

\begin{equation}
	\reallywidehat{\mu^0_c} = \frac{N_c(T)}{T}, \forall c \in \mathcal{C}
\end{equation}

\subsection{Pre-processing}
We now describe the pre-processing steps that we employ before estimating the latent network and the triggering kernel parameters.

\beginparagraph{\rupee1 Deal.} We simply filter out the promotional transactions that were triggered by the \rupee1 daily deals, and retain only the organic ones.

\beginparagraph{Re-seller.} We eliminate all the transactions by the purported re-sellers that are identified by an abnormally high purchase velocity. In particular, anyone who has purchased $10$ or more items (well above the limits set by the fair usage policy: for example, one cannot add more than $4$ units of $5l$ olive oil tin cans in one basket) from the same category within a sliding window of $7$ days.


\beginparagraph{Attribution.} Let $\mathcal{T}_{u,c}(t)$ denote the collection of timestamps when the user $u$ has purchased from category $c$, until time $t$. Given an ordered pair of categories $(c, ~c')$ where $c \neq c'$ and $c, c' \in \mathcal{C}$,  we define a matching $\mathcal{M}_{u,c,c'}(t)$ as a collection of all pairs of timestamps $(t_i, t_j)$ where $\forall ~t_i \in \mathcal{T}_{u,c}(t)$ we find $t_j \in \mathcal{T}_{u,c'}(t)$ such that $t_j = \argmin_{t_j \in \mathcal{A}_{u, c,c'}^{(i)}}{d_{u,c,c'}^{(i)}}$ where  $d_{u,c,c'}^{(i)} \coloneqq t_j - t_i$ and $\mathcal{A}_{u, c,c'}^{(i)} = \{t_j~|~ t_i < t_j <t_i + 10 ~\text{(days)}\}$.


\subsection{Triggering Kernel Estimation}
For each $u \in \mathcal{U}$ and $(c,c') \in \mathcal{C} \times \mathcal{C}$, we extract the average inter-purchase interval, $\overline{d_{u,c,c'}}$, from the matching $\mathcal{M}_{u,c,c'}(T)$ by taking a weighted average of $\big\{ d_{u,c,c'}^{(i)} \big\}_{i = 1}^{\mid \mathcal{T}_{u,c}(t) \mid}$, where the weights are inversely proportional to the logarithm ($\log_2(2 + \cdot)$, to be precise) of the number of other purchases that has happened within the span of $d_{u,c,c'}^{(i)}$.
To estimate the parameters of the ${\abs{\mathcal{C}} \choose 2}$ \textsc{Weibull} kernels, for each pair of distinct categories, $c \neq c'$, we fit the Weibull distribution to $\big\{ \overline{d_{u,c,c'}} \big\}_{u \in \mathcal{U}}$. Similarly, for each $c \in \mathcal{C}$, we estimate the parameters of the \textsc{MoW} kernel by fitting a mixture of $K$ (set to $5$) Weibull distributions to $\big\{ \overline{d_{u,c,c'}} \big\}_{u \in \mathcal{U}}$ with EM. The figure \ref{fig:mow_fit} shows the \textsc{MoW} kernel fit to inter purchase times of \textit{Sugar} category with $K = 5$.

\subsection{Latent Network Estimation}
We now elaborate on Lifted Markov estimator for the latent network. We begin with Markov Estimator(\MKV), which intuitively captures the probability of purchase in $c$ that immediately follows a purchase in $c'$:
\begin{equation}
	\reallywidehat{\beta^{MKV}_{c,c'}} = \frac{\sum_{u \in \mathcal{U}} \abs[\big]{\mathcal{M}_{u,c,c'}(T)} + \alpha}{\sum_{u \in \mathcal{U}} N_{u,c'}(T) + \abs[\big]{\mathcal{C}} \beta}
\end{equation}
where $\alpha$ and $\beta$ are smoothing parameters that we set to $3$ and $.1$, respectively.

\beginparagraph{Lifted Markov Estimator (\LMKV).} The Markov estimator tends to connect popular categories to the rest, which the \textit{lifted} Markov estimator attempts at correcting:

\begin{equation}
	\reallywidehat{\beta^{LMKV}_{c,c'}} = \frac{\reallywidehat{\beta^{MKV}_{c,c'}}}{\sfrac{N_c(T)}{\sum_{c \in \mathcal{C}} N_c(T)}}
\end{equation}

	
\begin{figure}
    \centering
    \includegraphics[width=0.95\columnwidth]{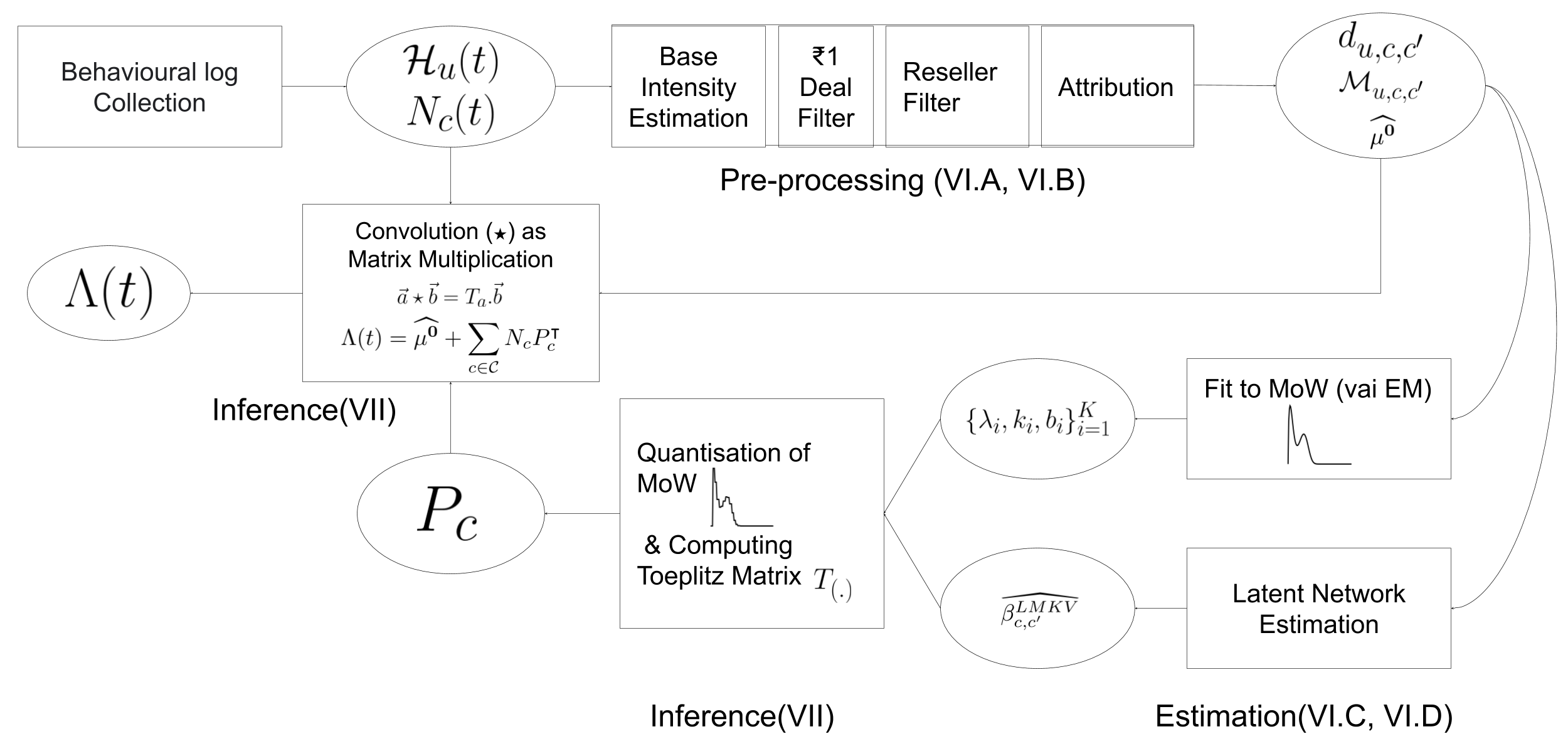}
    \caption{An overview of the estimation (Section \ref{Estimation}) and inference (Section \ref{Inference})
     algorithm. The behavioural log goes through pre-processing step before Likelihood Free estimation. Once the parameters are estimated, we scale the Inference employing Quantized MoW formulation and implementing convolution as Matrix Multiplication.}
    \label{fig:algo}
\end{figure}

\section{Inference} \label{Inference}
We are now in a position to illustrate the inference procedure for \A~that solely relies on matrix multiplication and addition operations.

\beginparagraph{Convolution as a Matrix Multiplication.} The result of the convolution operation between a pair of vectors, $a \in \Real^n$ and $b \in \Real^n$, is another vector, $a \star b \in \Real^{2n-1}$, which can be expressed as a matrix-vector product between, $T_a$ -- a Toeplitz matrix extracted from $a$ and the vector $b$ \cite{conv}.

\beginparagraph{Precomputes for Inference.} We pre-compute Toeplitz matrices required to implement convolution as Matrix Multiplication. Let $K_{c} \in \Real^{\mid \mathcal{C} \mid \times \mid t-1 \mid}$ denote the quantised triggering kernel levels for the $(0, t-1]$ period for category $c$ i.e., for every pair of category $c'$ and day $t'$, the corresponding matrix entry is $K_{c}(c', t') = \kappa_{c,c'}(t')$. Now for every category $c$ we compute the following matrix $P_{c} = (B_{c} \mathbbm{1}_{t-1}^{\intercal}) \odot K_{c}$ where $\mathbbm{1}_{a}$ denotes the vector of $1$'s of size $a$, $B_{c} \in \Real^{\mid \mathcal{C} \mid}$ with $B_{c}(c') = \beta_{c, c'}$(for every category $c'$) and $\odot$ denotes the Hadamard (element-wise) product.

\beginparagraph{Aggregation and inference.} We let $\Lambda(t) \in \Real^{\mid \mathcal{U} \mid \times \mid  \mathcal{C} \mid}$ denote the vector of intensities that \A~infers for all the users in $\mathcal{U}$ for day $t$. Furthermore, we let $N_{c} \in \Real^{\mid \mathcal{U} \mid \times t-1}$ denote the daily purchase counts of the all users on category $c$ until day $t-1$. With this notation, $\Lambda(t)$ can be computed as follows, where $\reallywidehat{\mathbf{\mu^0}} \in \Real^{\mid \mathcal{U} \mid \times \mid \mathcal{C} \mid}$ denote the matrix of estimated base intensities :
\begin{equation}\label{eq:inference}
	\Lambda(t) = \reallywidehat{\mathbf{\mu^0}} + \sum_{c \in \mathcal{C}} N_{c} P_{c}^{\intercal}
\end{equation}

\section{System Architecture} \label{System}

We now describe the design and implementation of the production system. The application shares a fleet of $150$ machines -- each provisioned with $20$ CPU cores, $53$ GB RAM, and $24$ TB SSD storage, connected via a $2$ Gbps network interconnect -- with several other throughput-sensitive production workloads. The resources are virtualized with containers.

\beginparagraph{Behavioural Log.} The Supermart client application sends a log of user interaction events to the server, which is relayed over a streaming topology to a central data lake. The event enrichment system performs a nightly lookup for the day's arrivals to obtain current category assignments (the product taxonomy is mutable) of the items and persists the resulting behavioural log, $\mathcal{H}_u(t), \forall u \in \mathcal{U}$, in a distributed file system, HDFS. This system processes $2.5$ TB of events daily in $15$ minutes with $400$ containers, each provisioned with $5$ GB of executor memory and $1$ CPU core.

\beginparagraph{Triggering Kernel \& Latent Network Estimation.} The triggering kernel and latent network estimation exploits the inherently parallel computations and is scheduled for nightly execution atop the behavioural log 
. The results 
, $P_c, \forall c \in \mathcal{C}$, are persisted into HDFS upon completion.

\begin{figure}
    \centering
    \includegraphics[width=0.95\columnwidth]{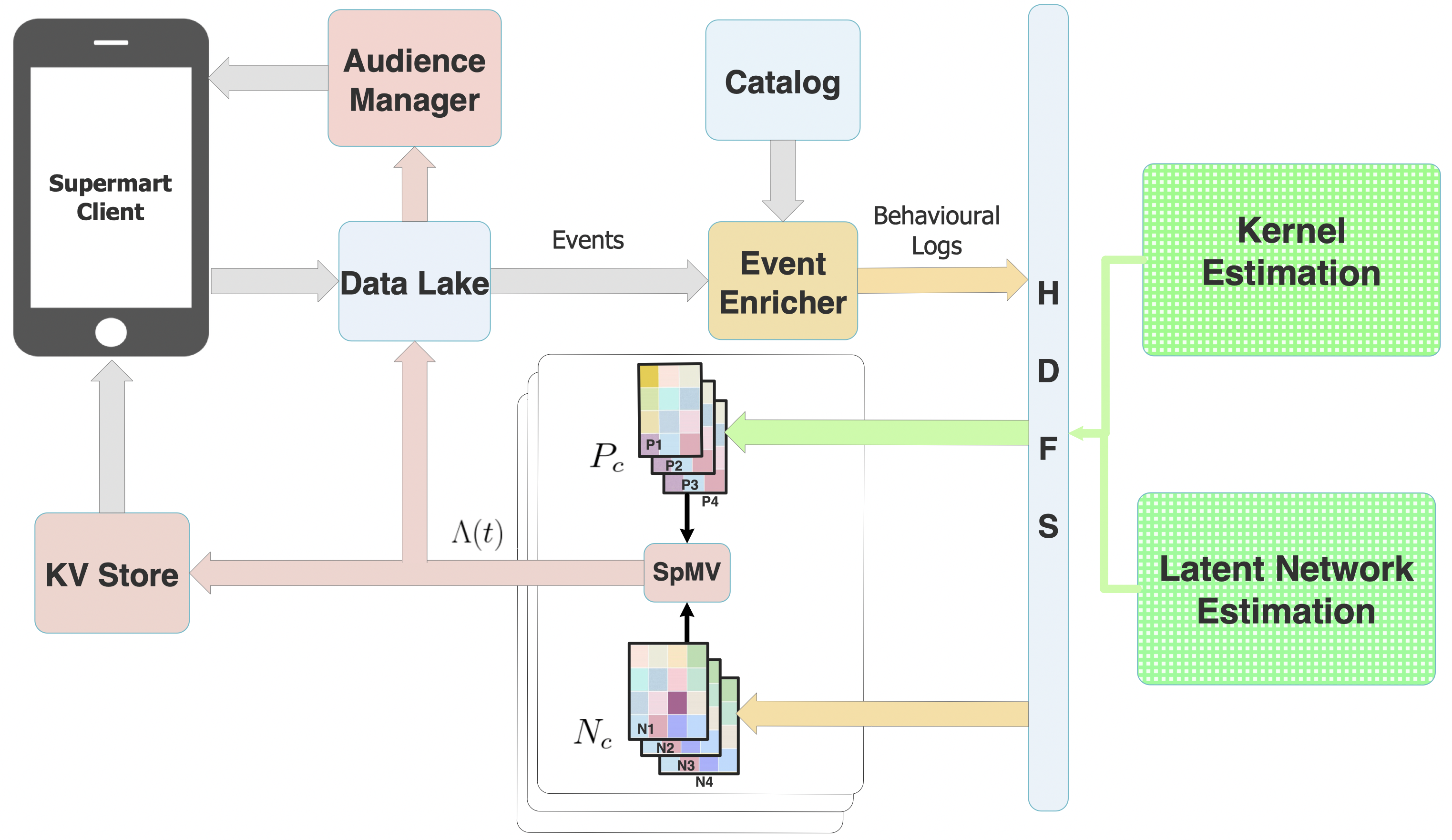}
    \caption{\A~system architecture (best viewed in colour): the colour on the boxes and arrows indicate data lineage. As an example, $P_c$ originates from the kernel estimation and the latent network estimation systems, whereas the lineage for $N_c$ is traced back to the event enrichment system.}
    \label{fig:System_Architecture}
\end{figure}

\beginparagraph{Aggregation.} The aggregation system maintains the event counts, $N_c, \forall c \in \mathcal{C}$, in an incremental fashion atop Apache Spark. The counts pertain to the past $\Delta = 180$ days. With $400$ containers, the aggregation step finishes in $10$ minutes and the aggregates are persisted with Avro compression.

\beginparagraph{Convolution.} Apache Spark makes native linear algebra libraries available to the application and allows us to optimise the implementation of Eq.~\ref{eq:inference}. We note that $P_c, \forall c \in \mathcal{C}$ can be cached in-memory ($\leq 0.5$ GB in size), allowing Spark to broadcast it and use native linear algebra library in each worker. Each worker operates on a slice of the $N_c$ matrix corresponding to a subset of the users, $U \subset \mathcal{U}$. With $400$ containers, this step finishes in $10$ minutes, thus bounding the end-to-end run time to less than half-an-hour.


\beginparagraph{Audience Manager.} The resulting $\Lambda(t)$ is persisted back to the data lake, where it becomes available for consumption by the \textit{Audience Manager} subsystem, a proprietary demand-side platform designed for the category managers and the advertisers for configuring target audiences. Note that we materialise several audiences (by sorting users in a descending order of $\Lambda_{u, c}(t)$ for the category of interest, $c$, and retaining the top-ranking users based on the reach requirement of the audience) and cache them in the audience manager for efficiency. Lastly, $\Lambda(t)$ is loaded into a key-value store and is used by the Flipkart Supermart client for several personalisation use-cases beyond precision merchandising.
\section{Experiments} \label{Experiments}
In this section, we show the efficacy of the proposed model through offline and online experiments. In particular, we conduct offline experiments on a targeting task - pick top-$k$ users who are interested in the category. We conduct online experiments on two audience creation tasks - homepage banner campaigns and push notification campaigns.
\subsection{Offline Experiments}

\beginparagraph{Dataset.} The dataset contains a large-scale sample from the Flipkart Supermart behavioural logs. In particular, it contains $27$ million purchases by $1.6$ million consumers across $90$ categories (grocery and FMCG) over a period of $15$ months. We only considered repeat customers: i.e., consumers with a minimum of $2$ purchases.

\beginparagraph{Train-Test Split.} A time based split into train and test is employed, where data from the last $2$ months is used for testing and the rest of the $13$ months of data is used for training. The construction of the split ensures that a user found in the test set will also be in the train set. The test period contains $15\%$ of the total purchases. And $18 \%$ of the total users have made purchases in the test period.

\beginparagraph{Test Protocol.}  The test data is split into seven chronological segments, each of nine days length. The hyper-parameter tuning is carried out using cross-validation on the training set. For a test segment $s$, we use all the purchases up until the beginning of $s$ (including the training set) for learning the model parameters.

\beginparagraph{Baselines.} We consider the following baselines, each of which can be considered as an instance of a temporal point process:
\begin{itemize}
    \item \Tophist: A simple baseline that captures users with most purchases, and where $\lambda^*_{TOP}(t) = N_{u, c}(t)$.
    \item \Toplatest: A variation of {\Tophist} which considers purchases in the last 45 days, and where $\lambda^*_{TOP}(t) = N_{u, c}(t) - N_{u, c}(t-45)$. 
    \item Matrix Factorization (\MF)~\cite{imf} : It has intensity $\lambda^*_{MF}(t) = \mathbf{u_t}^T \mathbf{c_t}$.
    \item \TIPAS~\cite{jure}: A state-of-the-art method in temporal point process literature for modelling the next action and it's timing.
    \item \BuyItAgain~\cite{rahul}: It uses a Poisson process to predict the next time of an action conditioning on the action.
\end{itemize}
where $\mathbf{u_t}$ and $\mathbf{c_t}$ are the latent vectors computed by carrying out {\MF} of $N(t) \in \Real^{|\mathcal{U}| \times |\mathcal{C}|}$, where a cell $N(t)[u, c] = N_{u, c}(t)$. We overload the notation of $N_{u, c}(t)$ to denote counts of any activities like clicks (for banner campaign) and number of notifications read (for push notification campaign) apart from orders. The {\Toplatest} baseline is considered to capture the inter purchase interval time, which is observed to be less than $45$ days for most categories. We note that baselines are comprehensive in nature, covering all the kernel design choices made in the literature and moreover {\TIPAS} has been compared against all the state of the art methods including the deep learning based point process.

\beginparagraph{Audience Creation.} For each category $c$, all the baselines provide a list of users ranked in decreasing order by $\lambda^*_{\cdot, c}(t)$. The target audience for a category $c$ is created by picking top $r_c$ ranked users from the list. The $r_c$ is termed as \textit{reach}, the size of the target audience.

\beginparagraph{Metrics.} Given a category $c$, reach $r_c$ and a test segment of length $\delta$, we begin by defining $\mathcal{U}_{c} = \{u \mid N_{u, c}(t + \delta) - N_{u, c}(t) > 0 \}$ - the set of users who have purchased from category $c$ in the test segment.
We now define Precision ($P_c$) and Recall ($R_c$) as follows:
\begin{equation}
    P_c = \frac{ \sum_{u \in \mathcal{U}_{c}} \mathbbm{1}_{\mathcal{A}_{c}}(u)}{r_c} \hspace{0.3in} R_c = \frac{\sum_{u \in \mathcal{U}_{c}} \mathbbm{1}_{\mathcal{A}_{c}}(u)}{|\mathcal{U}_{c}|}
\end{equation}
where $\mathcal{A}_{c} = \{u \mid \text{rank}(\lambda_{u, c}(t)) \le r_c \} $ and $\text{rank}(\cdot)$ provides the rank of the user $u$ over the decreasing ordered list of $\lambda_{u, c}(t)$ for category $c$. For offline experiments, we set $r_c = k*p_c$ where $k$ is an integer and $p_c$ is the average number of purchases in category $c$ in $\delta$ days. $p_c$ is calculated by partitioning the entire train set into several segments of size $\delta$, and averaging purchase counts. We observed that for the most popular category $p_c \approx 40$k, while average $p_c$ across all categories is $6$k. We reports average $P@k$ and $R@k$ values across categories, where $P@k=\frac{1}{|\mathcal{C}|}\sum_{c \in \mathcal{C}}P_c$ and $r_c=k*p_c$.

\beginparagraph{Results.} Table \ref{table:EntireOffline} shows results for different methods (refer to the ‘All’ Section). {\TIPAS} performs poorly compared to the other methods. {\A} outperforms all the other methods for values of $k > 5$. 

\begin{table}
\centering
\renewcommand{\arraystretch}{1.0}
\begin{adjustbox}{width=1.05\columnwidth,center}
\begin{tabular}{@{\hspace{\tabcolsep}} clcccccccc @{\hspace{\tabcolsep}}}
\toprule
\multirow{2}{*}{Cohort} & \multirow{2}{*}{Algorithm} & \multicolumn{4}{c}{Precision (in \%)} & \multicolumn{4}{c}{Recall (in \%)} \\
\cmidrule(lr){3-6} \cmidrule(lr){7-10}
 & & $@5$ & $@10$ & $@20$ & $@40$ & $@5$ & $@10$ & $@20$ & $@40$ \\
\midrule
\multirow{6}{*}{All} & \Tophist & $3.40 $ & $2.41 $ & $1.42 $ & $0.88 $ & $22.81 $ & $32.03 $ & $37.54 $ & $45.92 $ \\
& \Toplatest & $2.94 $ & $1.79 $ & $0.96 $ & $0.53 $ & $19.43 $ & $22.93 $ & $24.52 $ & $27.42 $ \\
& \MF & $2.58 $ & $1.95 $ & $1.40 $ & $0.95 $ & $17.46 $ & $26.02 $ & $36.72 $ & $48.88 $ \\

& \TIPAS & $2.17 $ & $1.63 $ & $1.08 $ & $0.67 $ & $14.89 $ & $21.80 $ & $28.40 $ & $34.93 $ \\
& \BuyItAgain & $\pmb{3.54}$ & $1.94 $ & $1.06 $ & $0.63 $ & $\pmb{23.66}$ & $25.93 $ & $28.40 $ & $33.34 $ \\
\cmidrule(lr){2-10}
& \A & $3.44 $ & $\pmb{2.52}$ & $\pmb{1.53}$ & $\pmb{0.95}$ & $23.04 $ & $\pmb{33.22}$ & $\pmb{40.10}$ & $\pmb{49.14}$ \\
\cmidrule(lr){2-10}
\midrule
\multirow{6}{*}{\NewToCategory} & \Tophist & $0.00 $ & $0.00 $ & $0.00 $ & $0.02 $ & $0.00 $ & $0.00 $ & $0.02 $ & $1.89 $ \\
& \Toplatest & $0.06 $ & $0.12 $ & $0.08 $ & $0.07 $ & $0.36 $ & $2.07 $ & $4.12 $ & $8.31 $ \\

& \MF & $0.93 $ & $0.80 $ & $\pmb{0.62} $ & $\pmb{0.43} $ & $3.86 $ & $8.15 $ & $16.68 $ & $\pmb{32.14} $ \\

& \TIPAS & $0.05 $ & $0.08 $ & $0.12 $ & $0.13 $ & $0.18 $ & $1.27 $ & $4.68 $ & $12.72 $ \\
& \BuyItAgain & $0.06 $ & $0.12 $ & $0.12 $ & $0.12 $ & $0.14 $ & $1.80 $ & $5.83 $ & $13.88 $ \\
\cmidrule(lr){2-10}
& \A & $\pmb{1.22}$ & $\pmb{1.00}$ & $0.59$ & $0.37$ & $\pmb{6.41}$ & $\pmb{12.99}$ & $\pmb{18.86}$ & $28.37$ \\
\cmidrule(lr){2-10}
\midrule
\multirow{6}{*}{\OldToCategory} 
& \Tophist & $3.40 $ & $2.41 $ & $1.46 $ & $1.43 $ & $44.94 $ & $\pmb{63.31}$ & $\pmb{74.56}$ & $\pmb{91.44}$ \\

& \Toplatest & $3.27 $ & $2.98 $ & $2.69 $ & $2.34 $ & $38.09 $ & $43.80 $ & $45.25 $ & $47.52 $ \\
& \MF & $3.95 $ & $3.29 $ & $2.72 $ & $2.30 $ & $27.81 $ & $39.92 $ & $53.25 $ & $65.05 $ \\

& \TIPAS & $2.31 $ & $2.01 $ & $1.80 $ & $1.70 $ & $29.43 $ & $43.45 $ & $55.08 $ & $62.03 $ \\
& \BuyItAgain & $4.03 $ & $3.70 $ & $\pmb{3.39}$ & $\pmb{2.99}$ & $\pmb{46.44}$ & $48.94 $ & $50.66 $ & $54.09 $ \\
\cmidrule(lr){2-10}
& \A & $\pmb{4.87}$ & $\pmb{3.90}$ & $2.80 $ & $2.21 $ & $40.5 $ & $55.87$ & $64.78$ & $72.60 $ \\
\cmidrule(lr){2-10}
\bottomrule
\end{tabular}
\end{adjustbox}
\vspace{0.1cm}
\caption{Results of Offline Experiments over the Entire dataset and cohorts New to Category (\NewToCategory) and Old to Category (\OldToCategory). The best performance in a given cohort and metric is highlighted.}
\label{table:EntireOffline}
\end{table}

\begin{figure*}[ht]
  \centering
  \begin{subfigure}[b]{0.31\textwidth}
    \includegraphics[width=1\textwidth]{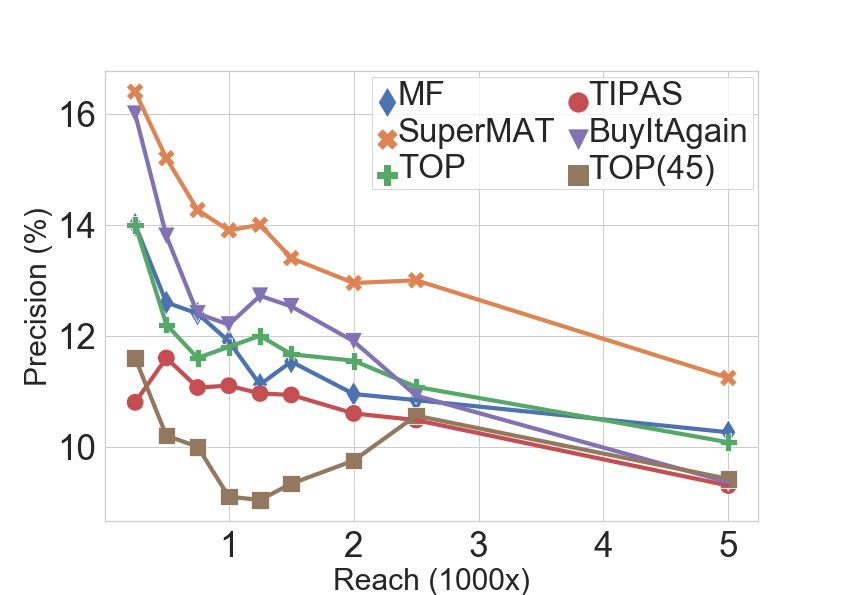}
    \caption{}
  \end{subfigure}
  \begin{subfigure}[b]{0.31\textwidth}
    \includegraphics[width=1\textwidth]{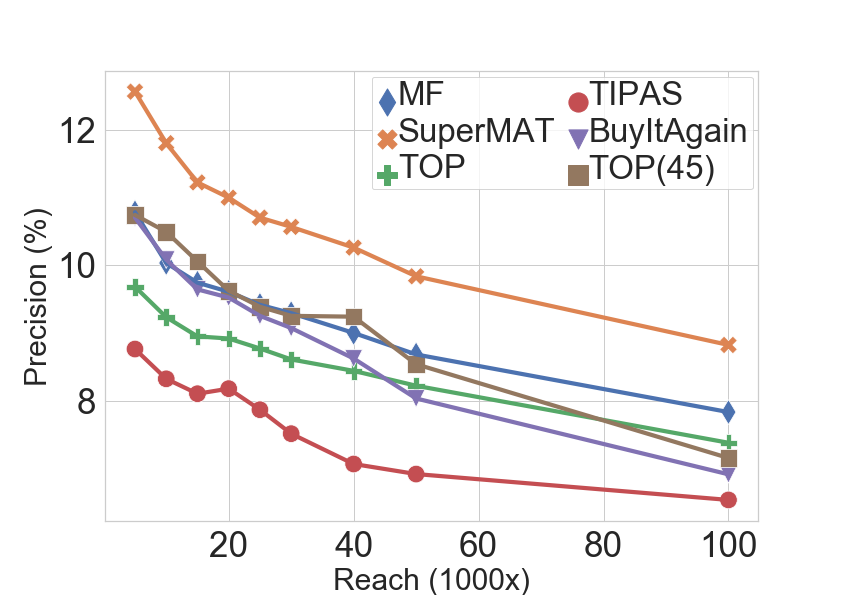}
    \caption{}
  \end{subfigure}
    \begin{subfigure}[b]{0.31\textwidth}
    \includegraphics[width=1\textwidth]{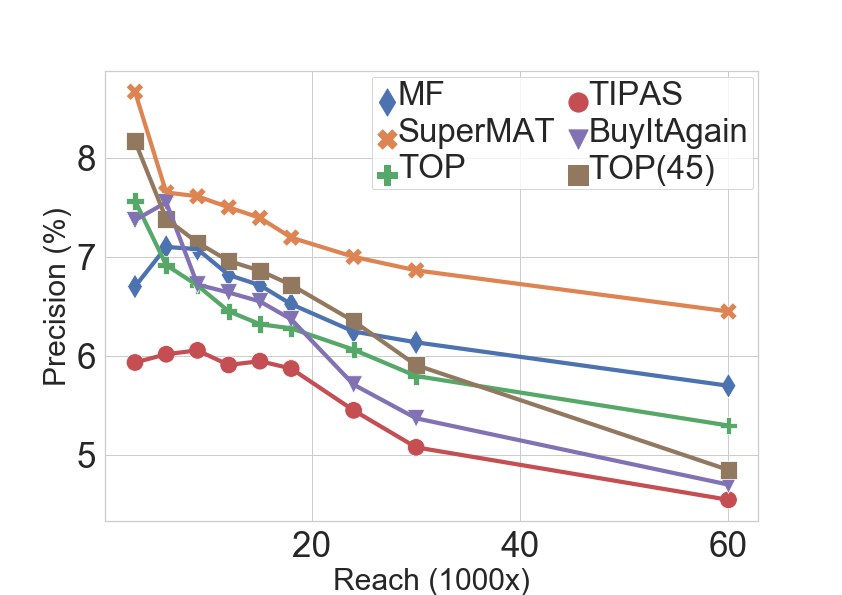}
    \caption{}
  \end{subfigure} \\
    \begin{subfigure}[b]{0.31\textwidth}
    \includegraphics[width=1\textwidth]{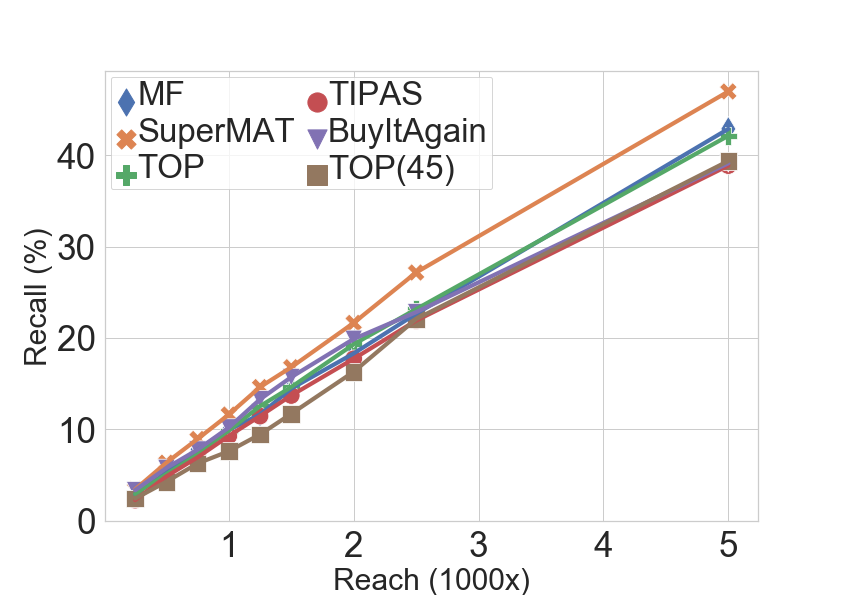}
    \caption{}
  \end{subfigure}
    \begin{subfigure}[b]{0.31\textwidth}
    \includegraphics[width=1\textwidth]{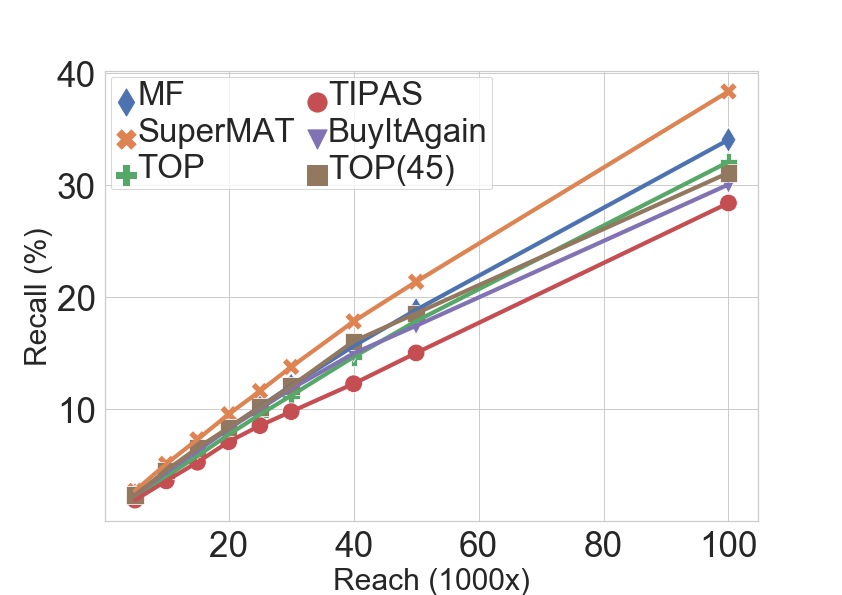}
    \caption{}
  \end{subfigure}
    \begin{subfigure}[b]{0.31\textwidth}
    \includegraphics[width=1\textwidth]{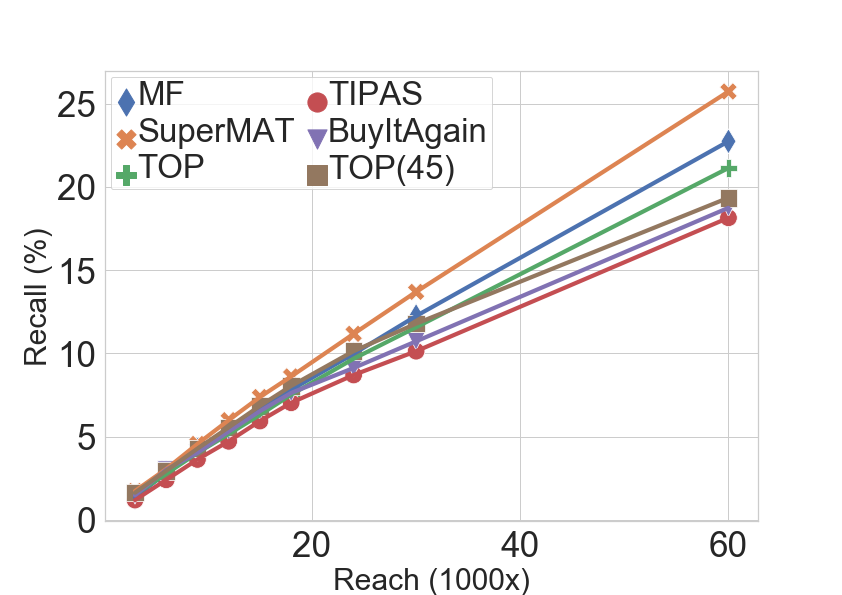}
    \caption{}
  \end{subfigure}
  \caption{Subplot a, b and c show precision comparison across methods for Atta, Detergent and Shampoo PN campaigns respectively; Similarly, subplot e, f and g show recall comparison across methods for the respective campaigns.}
  \label{fig:PNresults}
\end{figure*}

\begin{figure}[ht]
  \centering
    \begin{subfigure}[b]{0.31\textwidth}
    \includegraphics[width=1\textwidth]{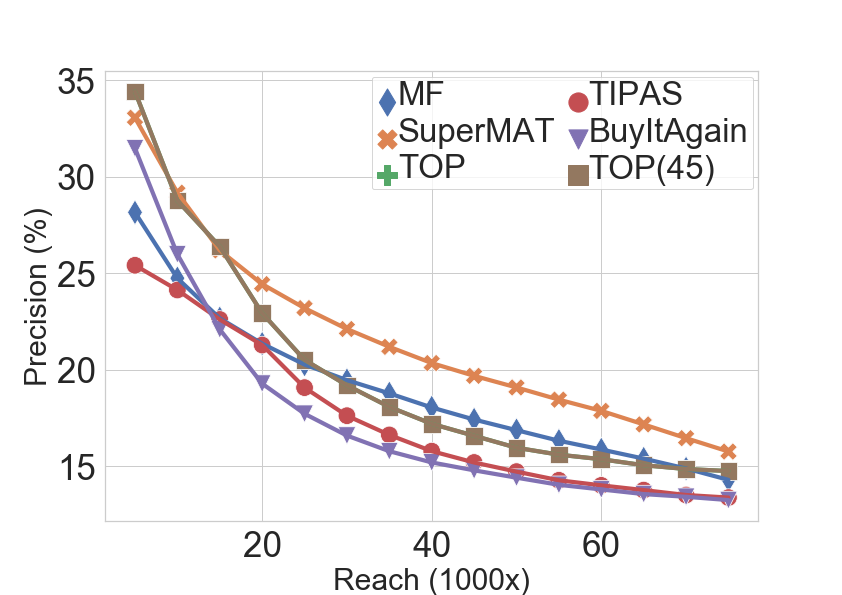}
    \caption{}
  \end{subfigure}\\
    \begin{subfigure}[b]{0.31\textwidth}
    \includegraphics[width=1\textwidth]{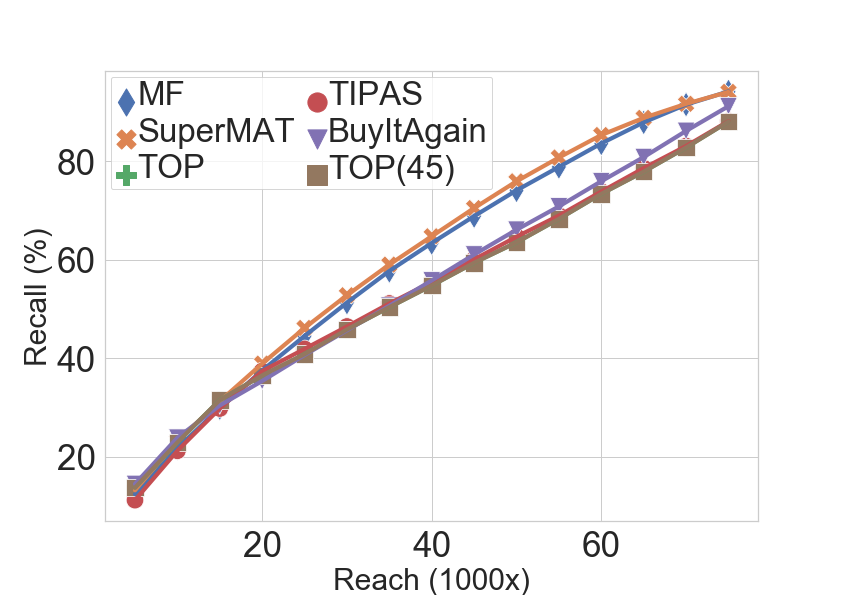}
    \caption{}
  \end{subfigure}
  \caption{Subplot a shows the precision comparison across methods and subplot b shows the recall comparison across methods for Atta Banner Campaign}
  \label{fig:bannerresults}
\end{figure}

\subsection{Cohort Analysis} Given the nascence and growth of the online consumable market, we expect a large number of new customers for a particular category. Particularly, we find that on an average $\sim 40\%$ of the purchases in a period of $\delta = 9$ days is done by such customers. This motivates us to define the following two cohorts and analyse them:
\begin{itemize}
    \item New to Category (\NewToCategory): Users who have not purchased from the category earlier.
    \item Old to Category (\OldToCategory): Users who have purchased from the category earlier.
\end{itemize} 
The {\NewToCategory} represents the growing segment of customers who explore our expanding list of categories  

\beginparagraph{Results on Cohorts.} Table \ref{table:EntireOffline} shows cohort based results. For lower reach, the performance of {\A} is clearly better than all other baselines on the {\NewToCategory} cohort. Similarly in the same cohort for higher reach, {\A} outperforms all the other baselines except {\MF}, where they are on par. Also, as expected {\Tophist} performs very poorly on this cohort, in-spite performing well on the entire dataset. The better performance of {\A} can be attributed to Latent Network Estimation which considers purchasers in the related categories. Although {\BuyItAgain} performs better than {\A} on {\OldToCategory} cohort, it has lower recall at higher values of $k$.





\subsection{Online Experiments}
\beginparagraph{Push Notification (PN) Campaigns.} Recurring Push Notifications, which reattempts to send the notification upon failure of the notification delivery to the user; were sent to a large segment of Supermart customers for three categories Atta (wheat flour), Detergent and Shampoo. The campaign statistics are summarized in Table \ref{table:PN_summary}. Each campaign was run roughly for a period of three days and the logged data of the PN campaign was replayed to all the competing baselines. The results in Figure \ref{fig:PNresults} show that {\A} performs better than the baselines with respect to both the metrics, precision and recall for different values of reach, the uplift obtained by {\A} against the best performing baseline is reported in Table \ref{table:PN_summary} for the maximum reach considered for each of the campaigns.
\begin{table}[htbp]
\centering
\renewcommand{\arraystretch}{1.2}
\begin{tabular}{@{\hspace{\tabcolsep}} lc @{\hspace{\tabcolsep}}}
\toprule
Campaign &  Uplift (in bps) \\
\midrule
Atta & $9.5$\\
Detergents & $12.8$\\
Shampoo & $13.18$\\
\bottomrule
\end{tabular}
\vspace{0.1cm}
\caption{Summary of the PN experiments showing number of users got PN, read the PN and finally the uplift seen(for maximum reach), for each of the categories}
\label{table:PN_summary}
\end{table}

\beginparagraph{Homepage Banner Campaigns.} The Supermart homepage was used to display campaign banners for Atta category to all the users landing on the homepage for a period of five days. A total of $\sim 100k$ Supermart customers were shown the campaign banner and $\sim 10k$ of these users clicked on the banner leading to a CTR of $13.11\%$. The impression and the click data of the campaign were replayed to all the competing baselines. The results in Figure \ref{fig:bannerresults} shows {\A}  performs better than the baselines for various values of reach. {\A} saw an uplift of $13.6$ bps compared to other baselines for a viable reach of $30k$.
\section{Conclusion} \label{Conclusion}
We have described the design and implementation of a precision merchandising system deployed at Flipkart Supermart since early $2019$, empowering hundreds of daily campaigns over an ever-growing customer base of ten million. With both offline and online experiments, we have demonstrated the superiority of the proposed algorithm over the state-of-the-art baseline. Incorporation of additional context -- e.g., family size and purchase quantity -- is left as a future work.

\bibliographystyle{IEEEtran}
\bibliography{grocery.bib}

\begin{thebibliography}{10}
\providecommand{\url}[1]{#1}
\csname url@samestyle\endcsname
\providecommand{\newblock}{\relax}
\providecommand{\bibinfo}[2]{#2}
\providecommand{\BIBentrySTDinterwordspacing}{\spaceskip=0pt\relax}
\providecommand{\BIBentryALTinterwordstretchfactor}{4}
\providecommand{\BIBentryALTinterwordspacing}{\spaceskip=\fontdimen2\font plus
\BIBentryALTinterwordstretchfactor\fontdimen3\font minus
  \fontdimen4\font\relax}
\providecommand{\BIBforeignlanguage}[2]{{%
\expandafter\ifx\csname l@#1\endcsname\relax
\typeout{** WARNING: IEEEtran.bst: No hyphenation pattern has been}%
\typeout{** loaded for the language `#1'. Using the pattern for}%
\typeout{** the default language instead.}%
\else
\language=\csname l@#1\endcsname
\fi
#2}}
\providecommand{\BIBdecl}{\relax}
\BIBdecl

\bibitem{mf}
Y.~Koren, R.~Bell, and C.~Volinsky, ``Matrix factorization techniques for
  recommender systems,'' \emph{Computer}, vol.~42, no.~8, p. 30–37, Aug.
  2009.

\bibitem{imf}
Y.~Hu, Y.~Koren, and C.~Volinsky, ``Collaborative filtering for implicit
  feedback datasets,'' in \emph{Data Mining, 2008. ICDM'08. Eighth IEEE
  International Conference on}.\hskip 1em plus 0.5em minus 0.4em\relax IEEE,
  2008, pp. 263--272.

\bibitem{fpmc}
S.~Rendle, C.~Freudenthaler, and L.~Schmidt-Thieme, ``Factorizing personalized
  markov chains for next-basket recommendation,'' in \emph{Proceedings of the
  19th International Conference on World Wide Web}, ser. WWW ’10.\hskip 1em
  plus 0.5em minus 0.4em\relax New York, NY, USA: Association for Computing
  Machinery, 2010, p. 811–820.

\bibitem{wei}
G.~Zhao, M.~L. Lee, W.~Hsu, and W.~Chen, ``Increasing temporal diversity with
  purchase intervals,'' in \emph{Proceedings of the 35th International ACM
  SIGIR Conference on Research and Development in Information Retrieval}, ser.
  SIGIR ’12.\hskip 1em plus 0.5em minus 0.4em\relax New York, NY, USA:
  Association for Computing Machinery, 2012, p. 165–174.

\bibitem{rahul}
R.~Bhagat, S.~Muralidharan, A.~Lobzhanidze, and S.~Vishwanath, ``Buy it again:
  Modeling repeat purchase recommendations,'' in \emph{Proceedings of the 24th
  ACM SIGKDD International Conference on Knowledge Discovery \& Data Mining},
  ser. KDD ’18.\hskip 1em plus 0.5em minus 0.4em\relax New York, NY, USA:
  Association for Computing Machinery, 2018, p. 62–70.

\bibitem{jure}
T.~Kurashima, T.~Althoff, and J.~Leskovec, ``Modeling interdependent and
  periodic real-world action sequences,'' in \emph{Proceedings of the 2018
  World Wide Web Conference}, ser. WWW ’18.\hskip 1em plus 0.5em minus
  0.4em\relax Republic and Canton of Geneva, CHE: International World Wide Web
  Conferences Steering Committee, 2018, p. 803–812.

\bibitem{alex}
H.~Jing and A.~J. Smola, ``Neural survival recommender,'' in \emph{Proceedings
  of the Tenth ACM International Conference on Web Search and Data Mining},
  ser. WSDM ’17.\hskip 1em plus 0.5em minus 0.4em\relax New York, NY, USA:
  Association for Computing Machinery, 2017, p. 515–524.

\bibitem{kitts}
B.~Kitts, D.~Freed, and M.~Vrieze, ``Cross-sell: A fast promotion-tunable
  customer-item recommendation method based on conditionally independent
  probabilities,'' in \emph{Proceedings of the Sixth ACM SIGKDD International
  Conference on Knowledge Discovery and Data Mining}, ser. KDD ’00.\hskip 1em
  plus 0.5em minus 0.4em\relax New York, NY, USA: Association for Computing
  Machinery, 2000, p. 437–446.

\bibitem{marketingbudget}
K.~Zhao, J.~Hua, L.~Yan, Q.~Zhang, H.~Xu, and C.~Yang, ``A unified framework
  for marketing budget allocation,'' 2019.

\bibitem{precisionmarketing}
W.~Y. Zou, S.~Du, J.~Lee, and J.~Pedersen, ``Heterogeneous causal learning for
  effectiveness optimization in user marketing,'' 2020.

\bibitem{tpp}
D.~R. Cox and V.~Isham, \emph{Point processes}.\hskip 1em plus 0.5em minus
  0.4em\relax London ; New York : Chapman and Hall, 1980.

\bibitem{scott}
A.~Sharma, R.~E. Johnson, F.~Engert, and S.~W. Linderman, ``Point process
  latent variable models of larval zebrafish behavior,'' in \emph{Proceedings
  of the 32nd International Conference on Neural Information Processing
  Systems}, ser. NIPS’18.\hskip 1em plus 0.5em minus 0.4em\relax Red Hook,
  NY, USA: Curran Associates Inc., 2018, p. 10942–10953.

\bibitem{chetlur2014cudnn}
S.~Chetlur, C.~Woolley, P.~Vandermersch, J.~Cohen, J.~Tran, B.~Catanzaro, and
  E.~Shelhamer, ``cudnn: Efficient primitives for deep learning,'' 2014.

\bibitem{lecture}
J.~G. Rasmussen, ``Lecture notes: Temporal point processes and the conditional
  intensity function,'' 2018.

\bibitem{leman}
E.~Manzoor and L.~Akoglu, ``Rush! targeted time-limited coupons via purchase
  forecasts,'' in \emph{Proceedings of the 23rd ACM SIGKDD International
  Conference on Knowledge Discovery and Data Mining}, ser. KDD ’17.\hskip 1em
  plus 0.5em minus 0.4em\relax New York, NY, USA: Association for Computing
  Machinery, 2017, p. 1923–1931.

\bibitem{conv}
K.~Pavel and S.~David, \emph{Algorithms for Efficient Computation of
  Convolution. Ch.8 of Design and Architectures for Digital Signal
  Processing}.\hskip 1em plus 0.5em minus 0.4em\relax IntechOpen, 2013.

\end{thebibliography}




\end{document}